\documentclass[runningheads]{llncs}
\usepackage[T1]{fontenc}
\usepackage{graphicx}
\usepackage{xcolor}
\usepackage{booktabs}
\usepackage{subfigure}
\usepackage{caption}
\usepackage{subcaption}
\usepackage{tabularx}
\usepackage{array}
\usepackage{courier}
\usepackage{amsfonts}
\usepackage{amsmath}
\usepackage{multirow}

\newcommand{\Dcal}{\mathcal{D}}
\newcommand{\Acal}{\mathcal{A}}
\newcommand{\Scal}{\mathcal{S}}

\newcommand{\eot}{\langle\mathrm{eot}\rangle} 

\newcommand{\error}[1]{\textbf{\textcolor{red}{#1}}}
\newcommand{\errorblue}[1]{\textbf{\textcolor{blue}{#1}}}

\begin{document}

\title{Normalized vs Diplomatic Annotation: A Case Study of Automatic Information Extraction from Handwritten Uruguayan Birth Certificates}

\titlerunning{Normalized vs Diplomatic Annotation}

\author{
Natalia Bottaioli\inst{1,2,3} \orcidID{0009-0005-3606-908X} 
\and Solène Tarride\inst{4} \orcidID{0000-0001-6174-9865} 
\and Jérémy Anger\inst{1} \orcidID{0009-0007-3319-6037}
\and Seginus Mowlavi\inst{1} \orcidID{0009-0005-7774-9217} 
\and Marina Gardella \inst{5} \orcidID{0000-0003-2465-2014} 
\and Antoine Tadros \inst{1} \orcidID{0000-0002-7885-903X} 
\and Gabriele Facciolo\inst{1} \orcidID{0000-0002-8855-8513} 
\and Rafael Grompone von Gioi\inst{1} \orcidID{0000-0002-6309-7116}  
\and Christopher Kermorvant\inst{4} \orcidID{0000-0002-7508-4080} 
\and Jean-Michel Morel\inst{6} \orcidID{0000-0002-6108-897X}
\and Javier Preciozzi \inst{2,3} \orcidID{0000-0002-9287-4826}
}

\authorrunning{N. Bottaioli et al.}

\institute{
Université Paris-Saclay, ENS Paris-Saclay, CNRS, Centre Borelli, France \and
Facultad de Ingeniería, Universidad de la República, Montevideo, Uruguay \and
Digital Sense, Montevideo, Uruguay \and
TEKLIA, Paris, France \and
IMPA, Rio de Janeiro, Brasil \and
City University of Hong Kong, Hong Kong
}

\maketitle        

\setcounter{footnote}{0} 

\begin{abstract}

This study evaluates the recently proposed Document Attention Network (DAN) for extracting key-value information from Uruguayan birth certificates, handwritten in Spanish. We investigate two annotation strategies for automatically transcribing handwritten documents, fine-tuning DAN with minimal training data and annotation effort. Experiments were conducted on two datasets containing the same images (201 scans of birth certificates written by more than 15 different writers) but with different annotation methods. Our findings indicate that normalized annotation is more effective for fields that can be standardized, such as dates and places of birth, whereas diplomatic annotation performs much better for fields containing names and surnames, which can not be standardized.

\end{abstract}

\section{Introduction}

Civil Registration and Vital Statistics (CRVS) is defined by the United Nations as the ``continuous, permanent, compulsory and universal recording of the occurrence and characteristics of vital events of the population in accordance with the law''.\footnote{https://unstats.un.org/unsd/demographic-social/Standards-and-Methods/files/Handbooks/crvs/crvs-mgt-E.pdf} CRVS is strongly related to article 6 of the Universal Declaration of Human Rights, which states that everybody has the right to a legal identity.\footnote{https://www.undp.org/africa/blog/having-legal-identity-fundamental-human-rights} 
Having a birth certificate gives to individuals legal evidence of their name, place and date of birth, and parents' names (among other data). In most countries, CRVS are key to ensure the right to have a legal identity, and this is clearly reflected by the fact that 67 of the 232 Sustainable Development Goals\footnote{https://sdgs.un.org/es/goals} are related to CRVS.\footnote{https://getinthepicture.org} Yet, more than 1 billion people worldwide do not have a document that serves as a proof of legal identity.

Because most CRVS were created long before the advent of computers, the majority of their registries are paper-based handwritten documents. This huge amount of documents is only recently being scanned and integrated into digital systems. Belize, Jamaica 
and Perú 
are some of the countries in America that are nowadays in the process of scanning historical CRVS data. 

 \begin{figure}[t]
     \centering
     \subfigure[Left margin]{\includegraphics[width=0.49\textwidth]{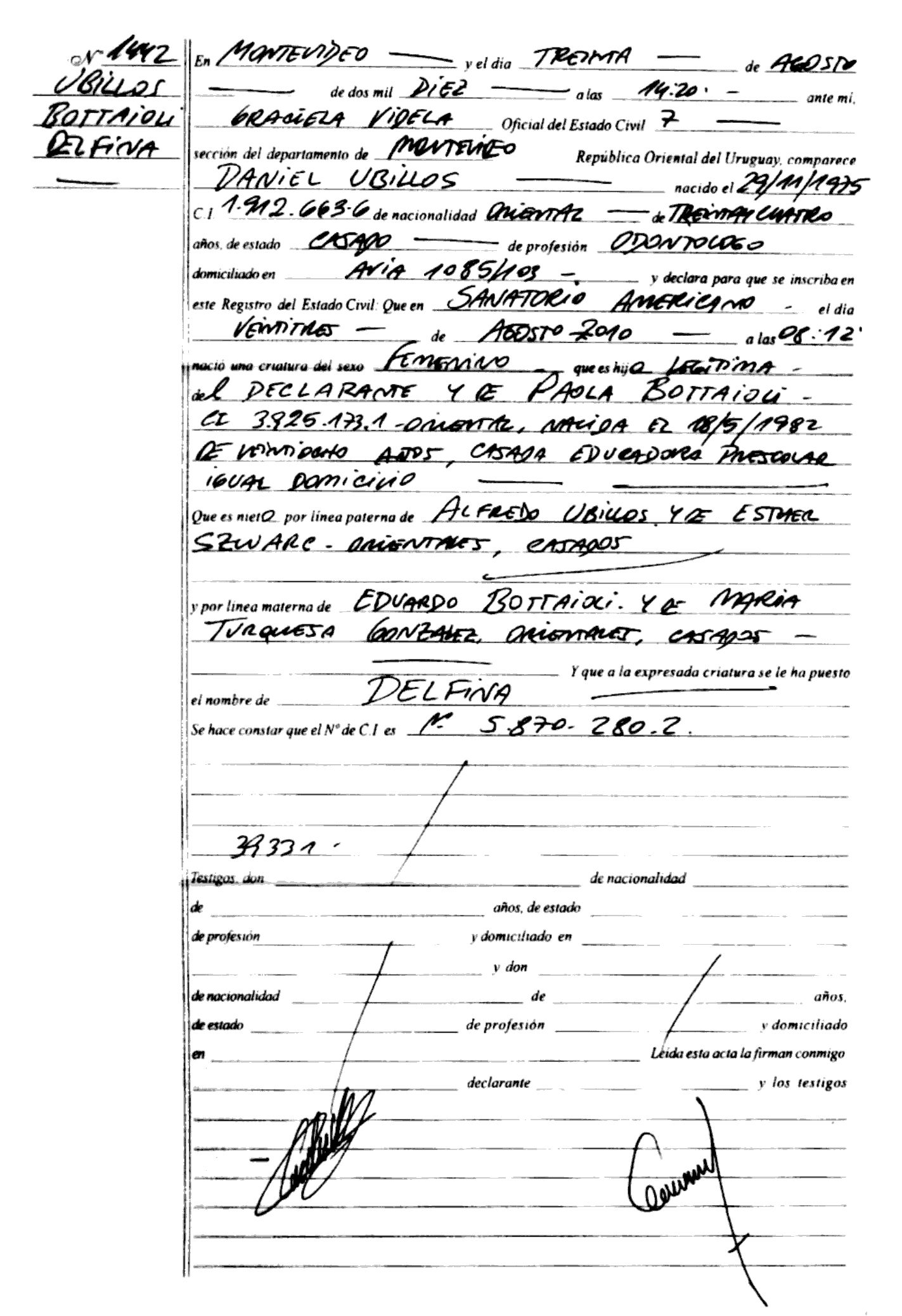}}
     \subfigure[Right margin]{\includegraphics[width=0.49\textwidth]{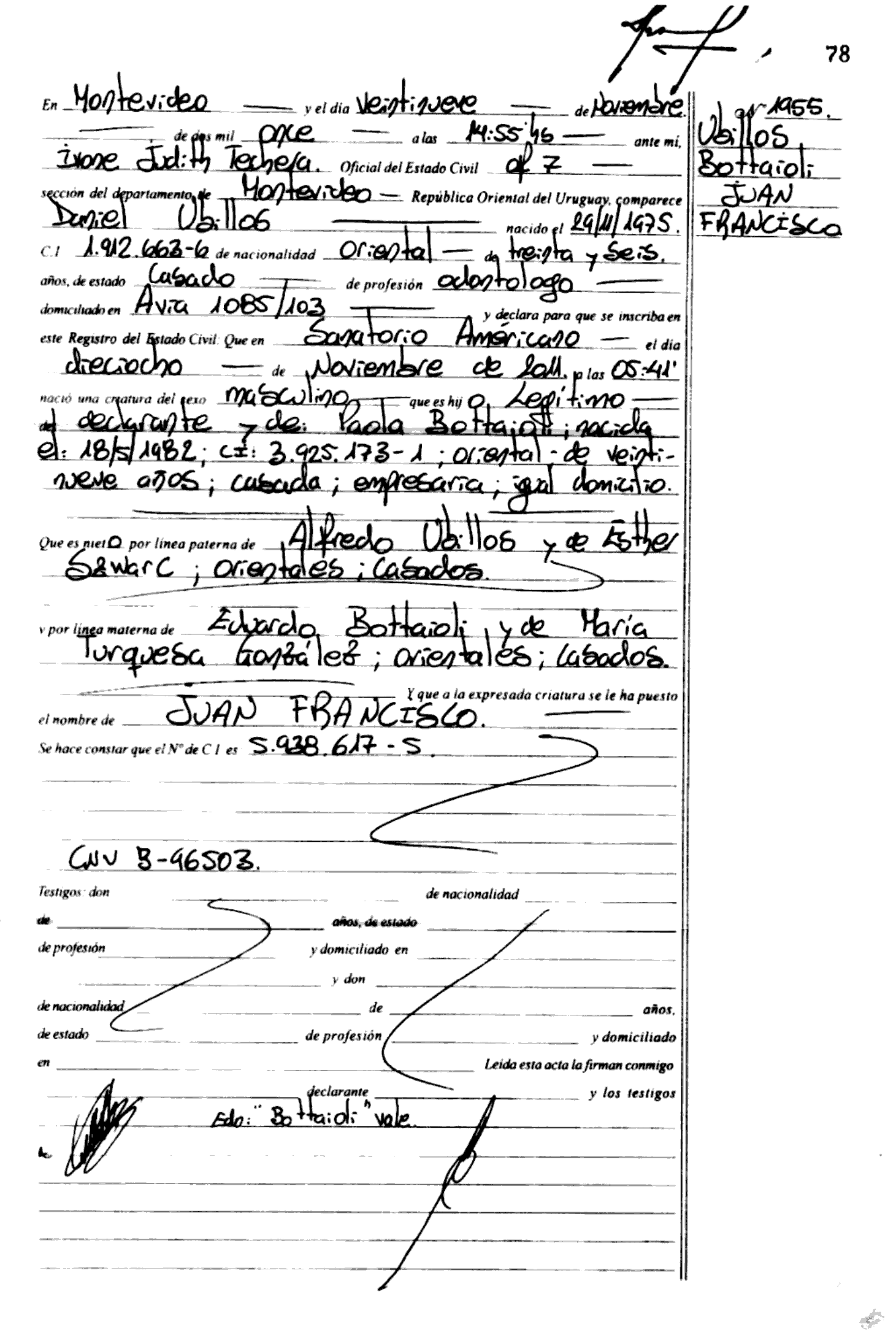}}
     \caption{Two scanned Uruguayan birth certificates, one with the margin on the left and the other one on the right. 
     }
     \label{fig:margen-types}
 \end{figure}

Here, we will focus on the case of Uruguay, where CRVS responsibility is held by a National Civil Registry Office, named the \textit{Dirección General del Registro de Estado Civil}.\footnote{https://www.gub.uy/ministerio-educacion-cultura/registro-civil} Initially named \textit{Registro de Estado Civil}, this office exists since 1879, when the government started keeping records of births, marriages, recognitions and deaths nationwide. In 2012, Uruguay's capital city, Montevideo, started registering birth certificates digitally, a process that was completed for the rest of the country only by 2022. Figure~\ref{fig:margen-types} shows two scanned birth certificates from Montevideo, one in which the margin is on the right and another one where the margin is on the left. It is worth mentioning that, due to the long period in which registries were generated manually, several different templates exist, and not all formats contain the same information. Even when the format is the same, the way of filling the data is not homogeneous in time. 

A digitization process has been gradually taking place in some departments but not yet nationwide: books have been scanned and, therefore, digital images of vital events have been generated. In order to make scanned images searchable and linked to other relevant registries, this digitization process is generally followed by a manual transcription of, at least, part of the data. This is a very demanding and time-consuming task, needless to say prone to errors. It is in this context that automatically processing these registries becomes of great interest.

The process of automatically recognizing the text present in an image requires both the interpretation of handwritten text (known as handwritten text recognition) and the layout interpretation of the scanned document (known as document layout analysis). This process generally requires the adaptation of the existing methods to a specific context. In~\cite{dan}, authors propose a Document Attention Network (DAN) for handwritten document recognition, where the adaptation process does not require to manually segment the different fields present in an image. In fact, only the scanned images and the fields to be transcribed with their corresponding annotations are required.

The objective of this work is twofold. First, we want to understand whether DAN can be adapted to work on the specific problem of handwritten document recognition for Uruguayan birth certificates, using very few data for the adaptation process. This analysis is relevant since DAN was trained for a different context (letters) and a different language (French). Second, we want to analyze two different annotation strategies. The first one consists of using \textbf{normalized} annotations, directly obtained from the Civil Registry Office. The second one uses \textbf{diplomatic} transcriptions (i.e. those that transcribe each and every character, respecting upper and lower cases, accents, etc). It is worth noting that in the first strategy, the data used to train the model was neither designed nor generated to train a deep learning model, while in the second one annotations were specifically generated for this task.

The rest of the article is organized as follows: Section~\ref{sec:related_work} describes the state of the art regarding handwritten document processing. 
Section~\ref{sec:model_description} reviews DAN, which is the model evaluated in this work. The Uruguayan birth certificates dataset is described in Section~\ref{sec:datasets}. Our two major experiments are presented in Section~\ref{sec:experiments}, while Section~\ref{sec:results} discusses the results. Conclusions are drawn in Section~\ref{sec:conclusions}, where future work is also described.

\section{Related work} \label{sec:related_work}

Handwritten document processing aims at producing some desired textual information from image inputs. It, therefore, sits at the intersection of computer vision (CV) and natural language processing (NLP). The difference in the modalities of these two fields has encouraged a sequential approach of the problem. The typical document processing pipeline involves one or several steps which perform transcription, i.e. mapping of the input image to a faithful textual output, followed by a natural language understanding step addressing a downstream task, e.g. key information extraction or named entity recognition. The large improvements brought by deep learning in CV~\cite{alexnet} and NLP~\cite{word2vec}, while strengthening the different steps of the sequential document processing pipeline, have also recently enabled integrated approaches. In the latter, algorithms perform the desired downstream task directly from the image input in an end-to-end fashion.

The two aforementioned approaches, sequential and integrated, lie at both ends of a spectrum that we briefly describe here. The sequential approach decomposes the workflow into a pipeline of simple subtasks, most often text line detection, handwritten text recognition (HTR), and finally the desired natural language understanding task. Current text line detection methods~\cite{dhsegment,docextractor} adapt the object detection literature, in particular CNNs~\cite{unet}, to the domain of document images. HTR methods~\cite{arora2019,puigcerver2017multidimensional} follow an encoder-decoder architecture, making use of decoders popularized in the speech-processing context such as hidden Markov models or recurrent neural networks with connectionist temporal classification~\cite{graves2009}. Finally, the NLP step is often carried out by a pre-trained transformer encoder model~\cite{bert} with simple decoder heads fine-tuned on the given task~\cite{flair,stanza}. Variations on this workflow include fusing and jointly training the text line detection and HTR steps~\cite{bluche2017,wigington2018}, or enriching the NLP step with the additional input modalities of layout and/or image besides text~\cite{LayoutLMv3,ErnieLayout,LayoutMask}. This added integration can bring specific benefits: for example, the former gains efficiency in the image embedding by using a single image encoder, and the latter augments the performance of the NLP step by allowing it to retain information from the  input image or the intermediate text line detection output. More recently, the potential of transformers as decoders of visual embeddings~\cite{trocr} has allowed for fully end-to-end models for document processing~\cite{dan,donut}.

An important part of the literature focuses on applying this body of research to settings arising from an industrial need~\cite{nion2013handwritten,balsac,esposalles,popp,simara,socface}. This type of work contributes back to the theoretical aspects in two key respects. First, it addresses the need for evaluation protocols able to eliminate the influence of benchmark-specific optimizations~\cite{cheplygina2023}, without expensive searches of the best training procedure for each method. Second, carrying out the entire document processing workflow enables the discovery of important insights beyond the issues of model design and training. For instance, a segmented approach benefits from the availability of powerful pre-trained algorithms~\cite{balsac} but runs the risk of errors accumulating across the steps~\cite{tarride2023key}. Hence, the choice of each algorithm needs to be guided by its input domain, e.g. the NLP module needs to be robust to ``recognition error noise'' in the input~\cite{abadie2022benchmark,monroc2022comprehensive}. Conversely,~\cite{balsac,petitpierre2023end} highlight the influence of the training dataset creation philosophy on the performance of HTR models and, thus, of sequential pipelines. Indeed, not only do annotations need to contain full transcriptions (even if a downstream key information extraction task only concerns a subset of that information), but transcriptions also need to stay faithful to the text (this is known as ``diplomatic annotation''). Diplomatic annotation becomes an issue when many database creators apply standardization to dates and names, expand abbreviations or introduce them, sometimes even without following consistent guidelines~\cite{catmus}. By contrast, integrated approaches can deal with partial annotation of training data~\cite{tarride2023key}. In addition, it has been shown that their multi-modality yields an increase in language modeling capabilities~\cite{tarride2022comparative}, which could translate into better flexibility with respect to unfaithfully annotated datasets.

\section{Model description}\label{sec:model_description}

For our experiments, we use DAN~\cite{dan,dangit}. We chose it for its off-the-shelf ability to be fine-tuned on data with a minimal amount of annotation effort, as it does not require bounding box-like annotation of text in the full image. Being an end-to-end architecture, it also allows us to study the influence of different annotation strategies in the context of an integrated approach to document processing.

The architecture of DAN follows the encoder-decoder design. The encoder is the fully convolutional network described in~\cite{van}, with the fixed 2D positional encoding of~\cite{singh2021full}. The decoder follows the original design of the auto-regressive transformer decoder~\cite{vaswani2017attention}, including the fixed 1D positional encoding. It generates tokens from a dataset-dependent dictionary $\Dcal=\Acal\cup\Scal\cup\{\eot\}$, where $\Acal$ is the set of characters, $\Scal$ is a set of semantic markers and $\eot$ is a special ``end of transcription'' token. The use of markers from $\Scal$ allows DAN to go beyond plain HTR, by segmenting the text output into layout elements or named-entities. It even allows increased flexibility: by adapting the ground-truth annotations with custom semantic markers and not necessarily including full transcriptions, one can train DAN to directly address other downstream natural language understanding tasks than those showcased by the authors, such as named entity recognition or key information extraction.

The model we use for fine-tuning has been trained in two phases. First, the network of the encoder and the prediction layer of the decoder have their weights initialized by transfer learning: a simple custom model is built using these modules for the task of text line transcription, and trained on a synthetic dataset of printed lines with the CTC loss. In the second phase, DAN is trained with the cross-entropy loss and teacher forcing on a combination of a real handwritten dataset and of synthetically generated printed documents, with the proportion of synthetic data gradually decreasing. The original authors ran the training on several datasets separately. In each case, synthetic documents are generated with the same layout classes as the dataset, with consistent positioning of the layout elements.

In the present work, we use the model trained in~\cite{dan} on the RIMES 2009~\cite{rimes2009} dataset. This dataset comprises grayscale images of French handwritten letters, with ground-truth annotation consisting in text regions with their layout class (one of ``sender'', ``recipient'', ``date and location'', ``subject'', ``opening'', ``body'' and ``PS and attachment'') and their text transcription. In order to train DAN, these annotations were automatically translated into sequences of tokens from $\Dcal_\mathrm{RIMES}=\Acal_\mathrm{RIMES}\cup\Scal_\mathrm{RIMES}\cup\{\eot\}$, with consistent rules for reading order determination. For our experiments, we use the open-source implementation of DAN~\cite{dangit}.

For our fine-tuning of this model, we need to use a token dictionary $\Dcal_\mathrm{FT}=\Acal_\mathrm{FT}\cup\Scal_\mathrm{FT}\cup\{\eot\}$ which is different from $\Dcal_\mathrm{RIMES}$. Indeed, our dataset is not in the same language (Spanish instead of French) thus, while the intersection of character sets $\Acal_\mathrm{FT}\cap\Acal_\mathrm{RIMES}$ is significant, both $\Acal_\mathrm{FT}\setminus\Acal_\mathrm{RIMES}$ and $\Acal_\mathrm{RIMES}\setminus\Acal_\mathrm{FT}$ are non-empty. Furthermore, our fine-tuning task of information extraction is encoded by a distinct set of markers than those for layout segmentation in RIMES, so $\Scal_\mathrm{FT}\cap\Scal_\mathrm{RIMES}=\emptyset$. As a consequence, we need to adapt the token embedding and probability prediction layers in DAN's decoder. Since their weight matrices have a clear interpretation, with individual tokens corresponding to their columns and rows respectively, this does not pose a significant challenge: we keep the trained weights corresponding to tokens in $\Dcal_\mathrm{FT}\cap\Dcal_\mathrm{RIMES}$, remove those corresponding to $\Dcal_\mathrm{RIMES}\setminus\Dcal_\mathrm{FT}$ and create new weights for $\Dcal_\mathrm{FT}\setminus\Dcal_\mathrm{RIMES}$.

\section{Datasets}\label{sec:datasets} 

\begin{figure}
    \centering
    \includegraphics[width=38em]{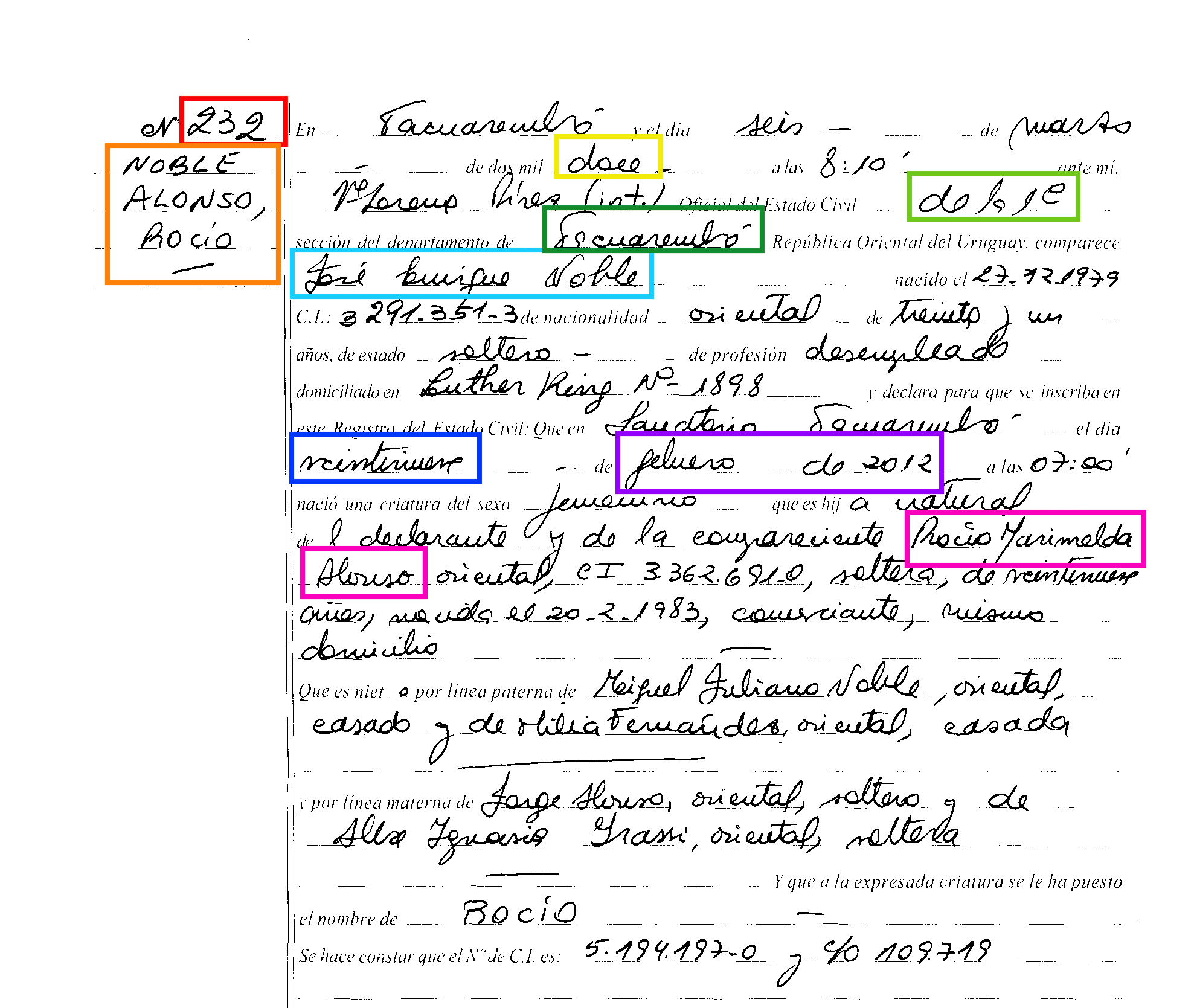}
\vspace{3mm}

\footnotesize
\begin{tabularx}{\textwidth}{ >{\hspace{5pt}}l<{\hspace{5pt}}  >{\hspace{5pt}}>{\ttfamily}X<{\hspace{5pt}}  >{\hspace{5pt}}>{\ttfamily}X<{\hspace{5pt}} }
\toprule
\multicolumn{1}{>{\hspace{5pt}}>{\bfseries}c<{\hspace{5pt}}}{\rule{0pt}{2.6ex}\textbf{Field}\rule[-1.2ex]{0pt}{0pt}} & \multicolumn{1}{>{\hspace{5pt}}>{\bfseries}c<{\hspace{5pt}}}{\rule{0pt}{2.6ex}\textbf{Normalized DAN}\rule[-1.2ex]{0pt}{0pt}} & \multicolumn{1}{>{\hspace{5pt}}>{\bfseries}c<{\hspace{5pt}}}{\rule{0pt}{2.6ex}\textbf{Diplomatic DAN}\rule[-1.2ex]{0pt}{0pt}} \\
\midrule
document number & 232 & 232 \\
enrollee's full name & Noble Alonso Rocio & NOBLE ALONSO ROCÍO \\
year of enrollment & doce & doce \\
jurisdiction & Primera & de la 1a \\
department & Tacuarembó & Tacuarembó \\ 
1st parent's name & José E. Noble & José Enrique Noble \\
date field 1 & veintinueve & veintinueve \\
date field 2 & febrero de 2012 & febrero de 2012 \\
2nd parent's name & Rocio M. Alonso & Rocío Marimelda Alonso \\
\bottomrule
\end{tabularx}
    \caption{Crop of the upper part of a birth certificate created by copying fields from several real birth certificates originally handwritten by the same public servant (and belonging to the dataset used in experiments), followed by a table showing two mentioned ways of annotating data. Annotated fields are manually highlighted in colors for visualization purposes only: document number (red), enrollee's full name (orange), year of enrollment (yellow), jurisdiction (light green), department (dark green), 1st parent's name (light blue), date field 1 (dark blue), date field 2 (violet), 2nd parent's name (pink). }
    \label{fig:crop_acta_anotada.png}
\end{figure}

\begin{figure*}[th]
   \centering
\begin{tabular}{cccc}
\includegraphics[width=0.235\textwidth]{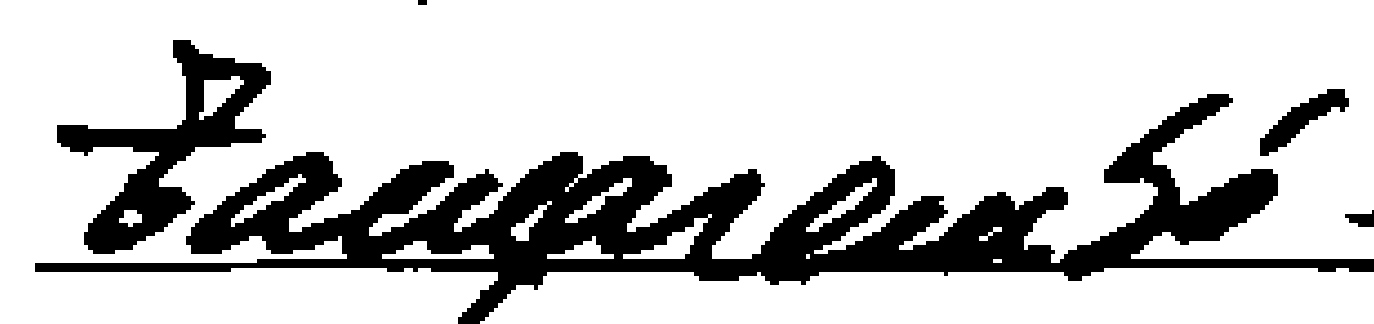}&
\includegraphics[width=0.235\textwidth]{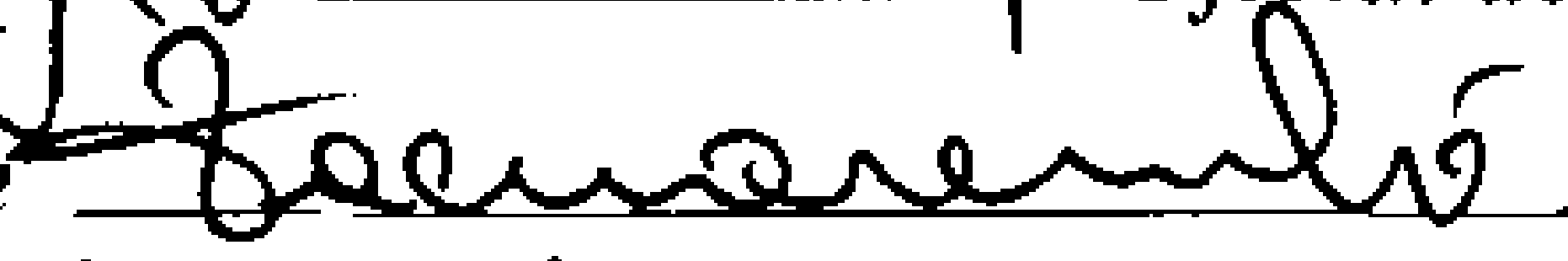}& 
\includegraphics[width=0.235\textwidth]{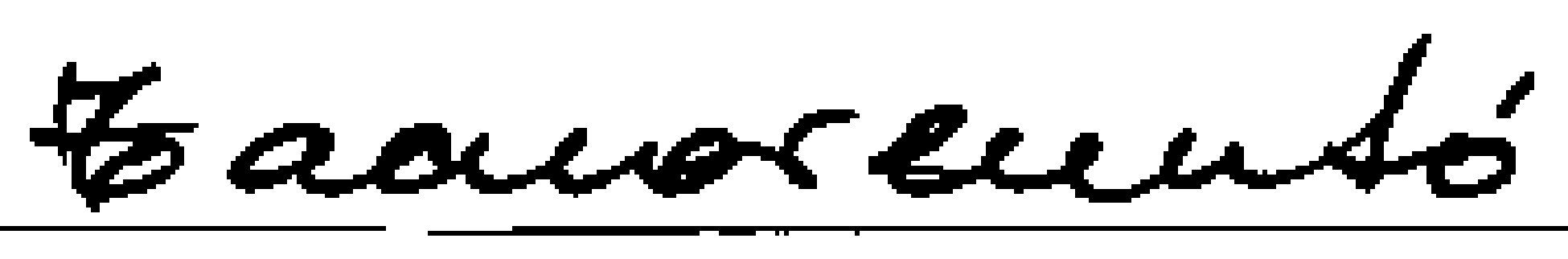}&
\includegraphics[width=0.235\textwidth]{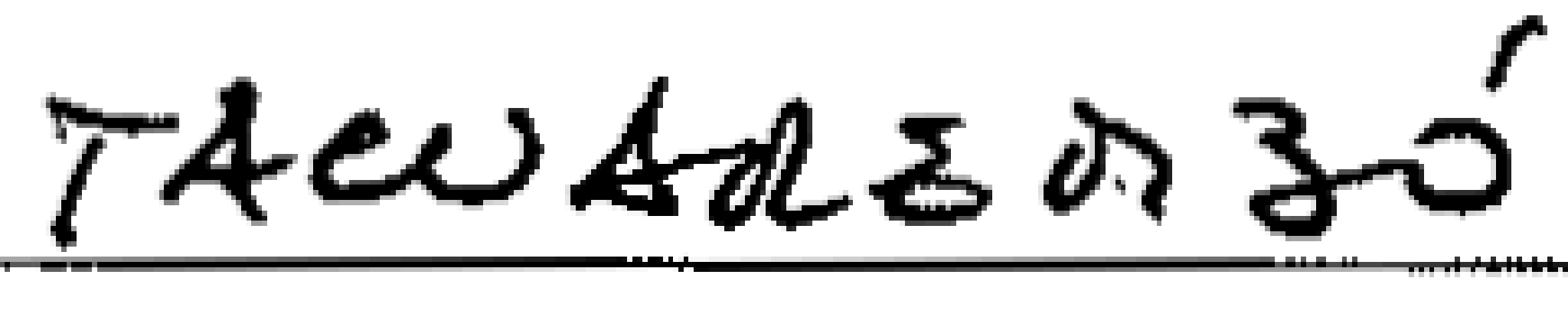}\\

\includegraphics[width=0.235\textwidth]{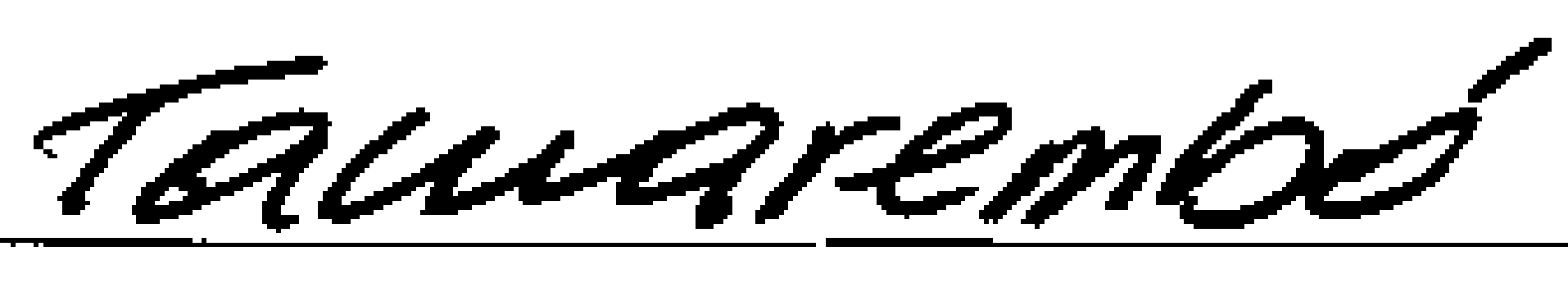}&
\includegraphics[width=0.235\textwidth]{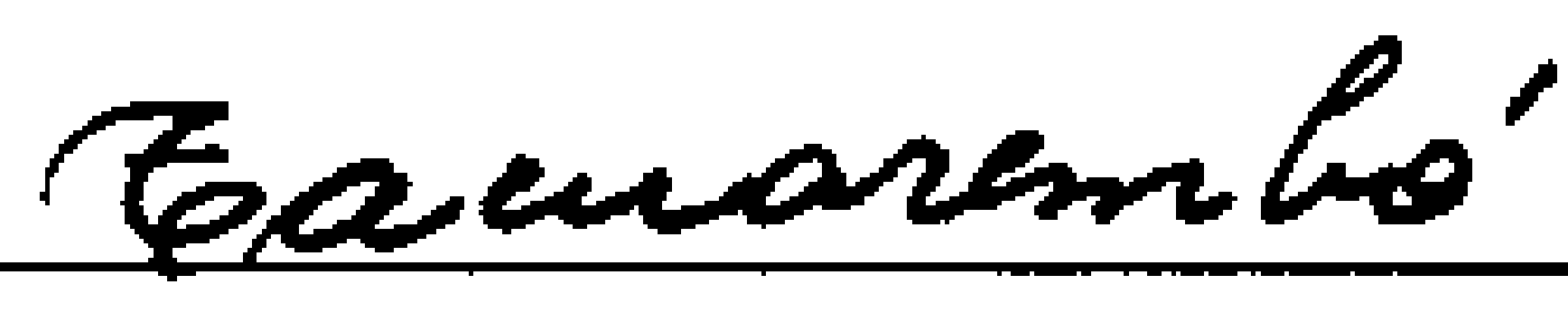}& 
\includegraphics[width=0.235\textwidth]{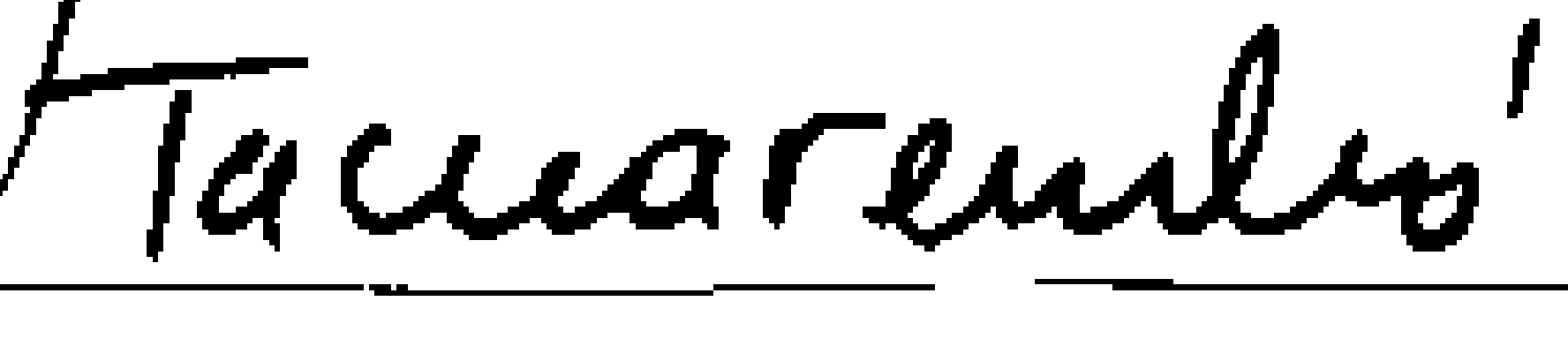}&
\includegraphics[width=0.235\textwidth]{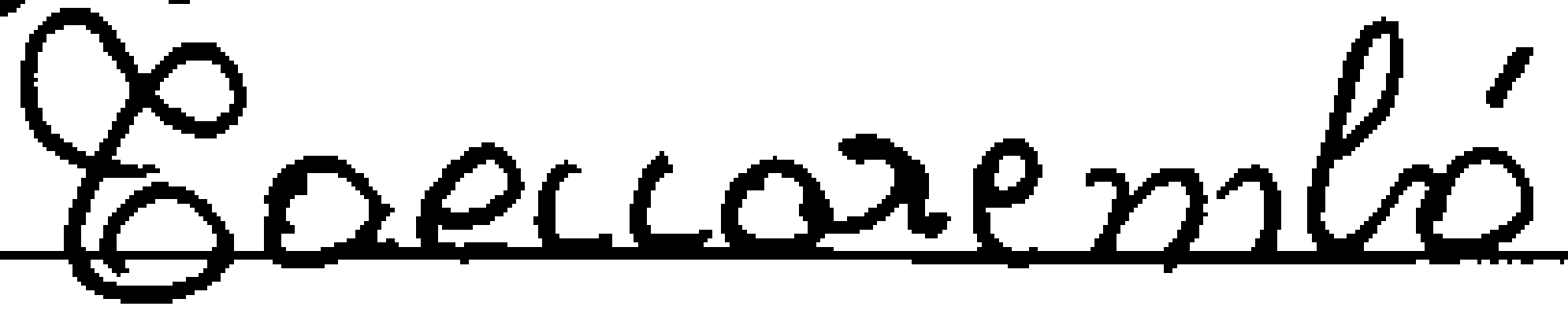}\\

\includegraphics[width=0.235\textwidth]{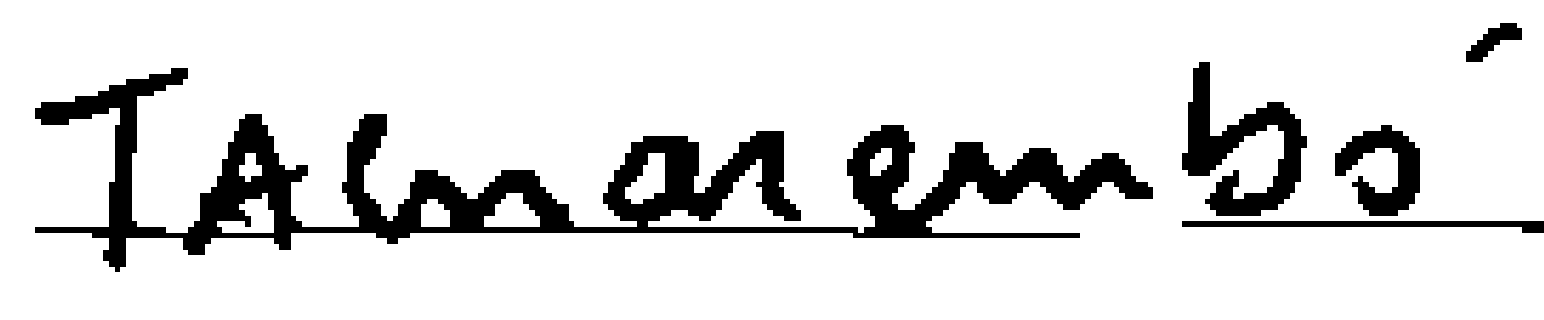}&
\includegraphics[width=0.235\textwidth]{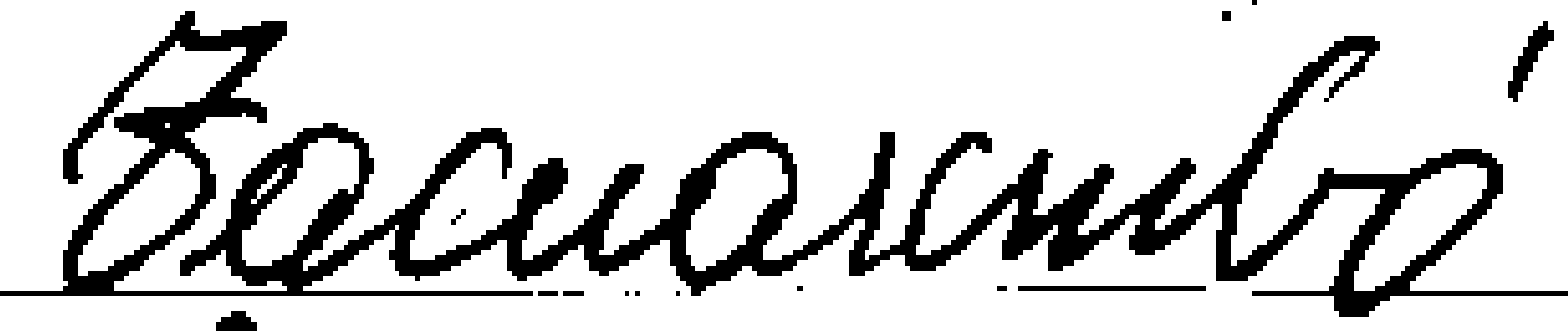}& 
\includegraphics[width=0.235\textwidth]{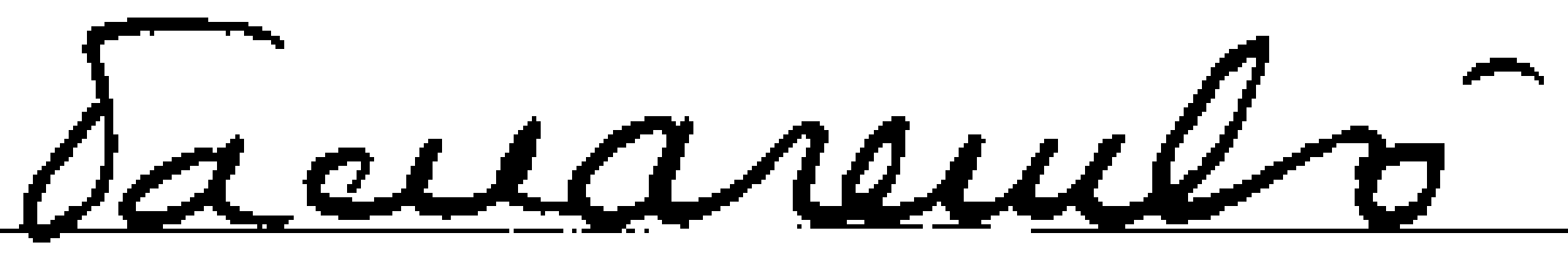}&
\includegraphics[width=0.235\textwidth]{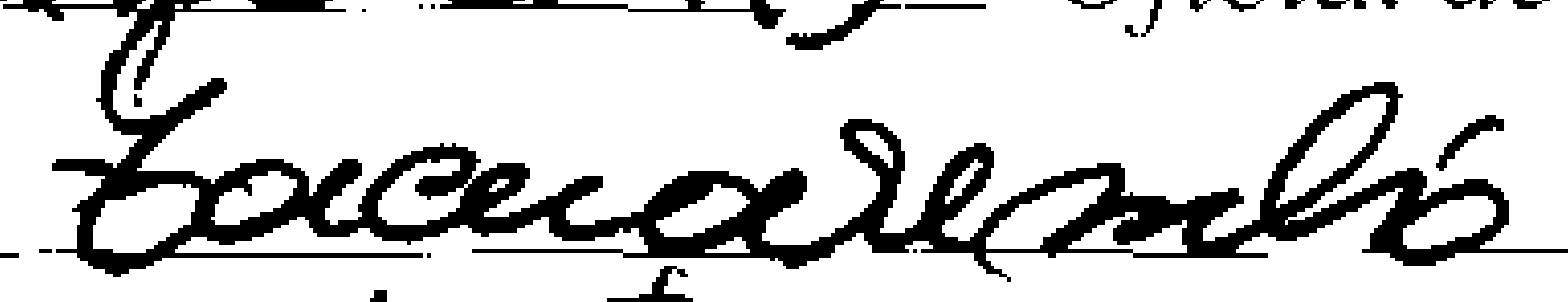}\\

\end{tabular}
\caption{Images of the word ``Tacuarembó'' as written by 12 different writers.}
    \label{fig:styles} 
\end{figure*}

For this work, we had access to a dataset composed of 201 birth certificates from 4 different years (2008, 2012, 2014 and 2016), handwritten in more than 12 writing styles (as can be seen in Figure~\ref{fig:styles}), with their corresponding transcribed text as kept in the computer system used in the local civil registry of Tacuarembó, Uruguay. Images are binarized and saved in tiff format, with a resolution of 200 dpi. The transcription of 7 fields present in the documents was provided in a \textsc{csv} file. These fields are: ``document number'', ``enrollee's full name'', ``year of the enrollment'', ``jurisdiction'', ``1st parent's full name'', ``date of birth'', and ``2nd parent's full name''. The field ``department'' was added to the dataset as a means of having one more annotation with zero annotation cost given that it is constant in all 201 documents. From this original dataset, we created two different sets of annotations. 

The first one was obtained by ordering the \textsc{csv} files containing the original annotations (as given by the civil registry of Tacuarembó) in the document's reading order, and converting the date, originally provided in the format ``YYYY-MM-DD'', into ``date field 1'' and ``date field 2''. The former contains the day of birth written in letters, while the latter contains the month written in letters followed by the word ``de'' followed by the year written in numbers (YYYY). (For instance, the date ``2014-05-31'' is annotated as ``treinta y uno'' in ``date field 1'' and ``mayo de 2014'' in the ``date field 2''.) This decision was made in order to avoid having to manually annotate the date in each of the documents. However, note that not all handwritten dates are written this way, as Table~\ref{tab:date-non-standard-format} shows.

\begin{table*}[htbp]
\addtolength{\tabcolsep}{4pt}
\begin{tabular}{lcc}

 & \multicolumn{2}{c}{\includegraphics[height=0.45cm]{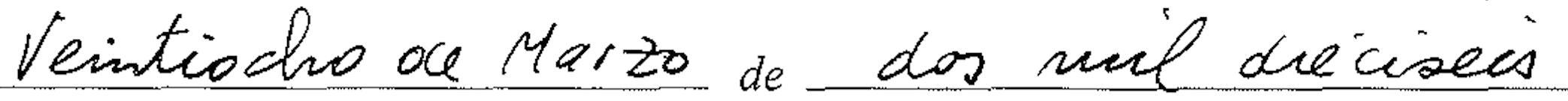}}\\
\cmidrule[\heavyrulewidth]{2-3}
 & date field 1 & date field 2 \\
\toprule
Normalized annotation & veintiocho & marzo de 2016\\
Normalized DAN inference & veintiocho & marzo de 2016\\
\midrule
Diplomatic annotation & Veintiocho de Marzo & dos mil dieciseis\\
Diplomatic DAN inference & Veintiocho de Marzo & dos mil dieciseis \\
\bottomrule
\end{tabular}
    \vspace{0.2 cm}
\caption{Birth certificate crop of a line where the date is handwritten in a way different from the most frequent one, together with annotations for ``date field 1'' and ``date field 2'' for both datasets. The inferred text by both models (Diplomatic DAN and Normalized DAN) is also included.}
\label{tab:date-non-standard-format}
\end{table*}

In the provided \textsc{csv} files, names are also transcribed in (possibly inconsistent) shortened ways. For example, ``Alberto Carlos Bustos'' can be transcribed as ``Alberto C. Bustos'' or as ``A. Carlos Bustos''. As a consequence, data kept in databases are not necessarily verbatim transcriptions of the handwritten text present in the documents. Although it could be ideal if this additional formatting was learned by the model to generate a formatted output (if desired), we decided to create another set of annotations, based on a verbatim transcription of the text (also known as ``diplomatic annotation''). 

This implied creating a new annotation set by: 1) transcribing all accents, when present; 2) respecting upper and lower case; 3) writing each of the two entities corresponding to the birth date in the way they appear in each document (dates can appear as ``veintiocho de marzo'' and ``dos mil dieciseis'' in some documents, as in the example shown in Table~\ref{tab:date-non-standard-format}, or as ``veintiocho'' and ``marzo de 2016'', which is the most usual way and the one used in the normalized annotation); 4) transcribing names and last names in full, as they appear in the documents (e.g. not abbreviating ``Maria Juana González'' as ``Maria J. González''). Generating the diplomatic transcription of all fields for the 201 documents took about 10 hours of manual work. 

Having these two sets of annotations (normalized and diplomatic) enabled us to perform two different experiments, which are described in the next section.

\section{Experiments}\label{sec:experiments} 

The main objective of the experiments was to understand the capabilities of DAN for transfer learning, in particular how good this process works in the context of structured birth certificates written in Spanish. We randomly divided the original dataset into train, validation and test in 80\% (161 images), 10\% (20 images) and 10\% (20 images), respectively, and used this same partition for all experiments. 

Fine-tuning was done using images of a fixed height of 1900 pixels, automatically resized in such a way that they preserved their aspect ratio. Original images were no smaller than 1800 pixels wide (therefore, they were, at the most, shrunk but never enlarged). The only pre-processing that images went through, before being used in the fine-tuning processes, was a subtle rotation in such a way that the side margin present in all images becomes vertical. 

The model features an encoder with 5 convolutional layers and a decoder with 8 transformer layers, each of size 256 with 4 attention heads, employing dropout, data augmentation, and label noise for enhanced performance. The maximum number of epochs is set to 2000, with a batch size of 5 and a learning rate of 0.0001.

We shall now describe our two experiments. The first one consisted in fine-tuning DAN using annotations that required very little pre-processing. We name the generated model \textbf{Normalized DAN}. The second one consisted in fine-tuning the same model using diplomatic (verbatim, character by character) annotations. We name the generated model \textbf{Diplomatic DAN}.

The reason why we started by training a first model using the annotation dataset obtained directly from the original \textsc{csv} files (computer-based annotations) is that the use of this kind of annotations is convenient for two reasons. First of all, there is no need for manual transcriptions, which is a costly and time-consuming process. Secondly, the desired result was that the model would output information in a normalized format that can be directly used by a computer system, without post-processing (as it is already formatted with the expected output). 

It is worth noting that in this first experiment the model must not only learn how to identify and extract the correct information but also how to format it to the desired output which is, a priori, a more complex task.

The second experiment was designed in order to test how well DAN would perform if fine-tuned with data that reflected exactly what is handwritten in the documents for the same set of fields (i.e. by using a diplomatic annotation).

\section{Results}\label{sec:results} 

Table \ref{tab:cer_wer-by_category} shows the Character Error Rate (CER) and the Word Error Rate (WER) for each field and each experiment. We have included two different values for each error measure: regular and normalized text. The first one is considering an exact, case and accent-sensitive comparison between text. For the normalized metric, we have considered uppercase and lowercase to be the same, and also accents were removed. We have included this analysis because, in several occasions, the differentiation of the casing or the presence of accents is not relevant. 

\begin{table}[t]
    \centering
    \setlength{\tabcolsep}{4pt}
    \begin{tabular}{l|rrrr|rrrr}
    \toprule
        \textbf{Field} & \multicolumn{4}{c|}{\textbf{Normalized DAN}} & \multicolumn{4}{c}{\textbf{Diplomatic DAN}}\\ 
         & \multicolumn{2}{c}{CER (\%)} & \multicolumn{2}{c|}{WER (\%)} 
         & \multicolumn{2}{c}{CER (\%)} & \multicolumn{2}{c}{WER (\%)} \\
         & reg. & norm. & reg. & norm. & reg. & norm. & reg. & norm. \\
       
        \midrule
        
        document number & 3.51 & - & 10.00 & - & 3.51 & - & 10.00 & - \\ 
                
        year of enrollment & 0.00  & \textbf{0.00}  & 0.00  & \textbf{0.00}  & 3.31     & 3.31  & 5.00  & 5.00 \\  
        
        jurisdiction & 0.00 & \textbf{0.00} & 0.00 & \textbf{0.00} & 0.92 & 0.92 & 3.33 & 3.33\\
        
        department & 0.00 & \textbf{0.00} & 0.00 & \textbf{0.00} & 4.50 & 0.00 & 5.00 & 0.00 \\

        date field 1 & 0.00 & \textbf{0.00} & 0.00 & \textbf{0.00} & 0.61 & 0.61 & 3.85 & 3.85 \\
        
        date field 2 & 0.00 & \textbf{0.00} & 0.00 & \textbf{0.00} & 0.35 & 0.00 & 1.67 & 0.00 \\

        enrollee's full name & 2.25 & 2.25 & 9.64 & 9.64 & 2.94 & \textbf{1.73} & 14.46 & \textbf{6.02} \\

        1st parent's name & 10.72  & 10.43  & 36.36  & 36.36  & 2.83  & \textbf{2.02}  & 12.86  & \textbf{10.00}  \\

        2nd parent's name & 15.81 & 15.81 & 35.19 & 35.19 & 7.17 & \textbf{5.70} & 28.36 & \textbf{20.90}\\

        \midrule
        
        global & 4.36  & 4.32  & 13.92  & 13.92  & 3.05  & \textbf{2.04}  & 11.27  & \textbf{7.51} \\
        
        \bottomrule

    \end{tabular}
    \vspace{0.2 cm}
        \caption{Character Error Rate and Word Error Rate for each field of birth certificates in the test set (composed of 20 birth certificate scans). Both metrics are computed on regular text (reg.) and normalized text (norm.), in which characters are changed to lowercase and accents are removed.}
    \label{tab:cer_wer-by_category}
\end{table}

\begin{table*}[t]
\addtolength{\tabcolsep}{7pt}
\begin{tabular}{lccc}
\toprule
Field & Annotation & Inference & Crops from documents\\
\toprule
\multirow{2}{*}{document number} & \ttfamily \error{9}0 & \ttfamily \error{2}0 &
\includegraphics[height=0.5cm]{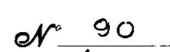}\\
& \ttfamily 115\error{4} & \ttfamily 115\error{6} & \includegraphics[height=0.5cm]{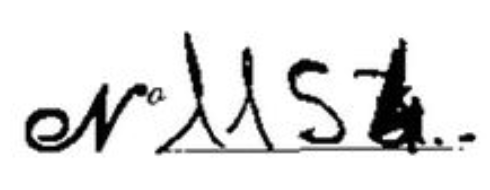}\\
\midrule
year of enrollment & \ttfamily catorce & \ttfamily c\error{o}ce\error{r} & \includegraphics[height=0.5cm]{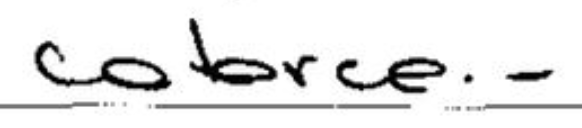}\\
\midrule
\multirow{2}{*}{jurisdiction} & \ttfamily de la 1\error{o} & \ttfamily de la 1\error{a} & \includegraphics[height=0.5cm]{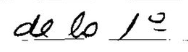}\\
& \ttfamily de la 1r\error{a} & \ttfamily de la 1r\error{\underline{ }}  & \includegraphics[height=0.5cm]{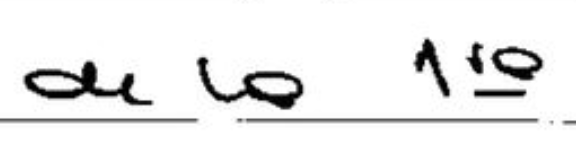}\\
\midrule
department & \ttfamily T\errorblue{ACUAREMBÓ} & \ttfamily T\errorblue{acuarembó} & \includegraphics[height=0.5cm]{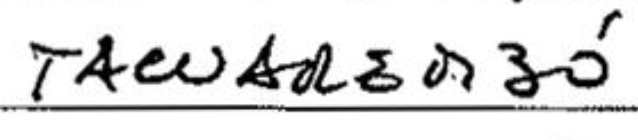}\\
\midrule
date field 1 & \ttfamily dieci\error{o}cho & \ttfamily dieci\error{s}cho & \includegraphics[height=0.5cm]{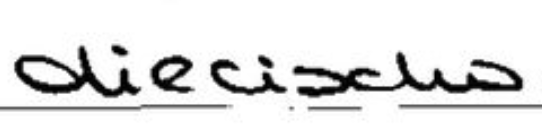}\\
\midrule
date field 2 & \ttfamily \errorblue{c}atorce & \ttfamily \errorblue{C}atorce & \includegraphics[height=0.5cm]{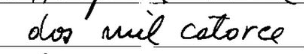}\\
\bottomrule
\end{tabular}
    \vspace{0.2 cm}
\caption{Errors made by Diplomatic DAN on normalizable fields. The only errors made by Normalized DAN in all normalizable fields are also shown in the table (they are the ones made in document number). Errors made when an uppercase letter instead of its corresponding lowercase letter (or vice-versa) are marked in blue. Errors made when actually substituting one character for another one or omitting a character are marked in red. No insertion errors were detected in these fields.}
\label{tab:errors_normalized_fields}
\end{table*}

A first observation of the quantitative results shown in Table~\ref{tab:cer_wer-by_category} is that both CER and WER values are similar to those obtained in~\cite{dan}, even though the context is different. Comparing the results of the two different models shows that \textbf{Diplomatic DAN} globally performs slightly better than \textbf{Normalized DAN}. Yet, a closer look at the results obtained for each field reveals that \textbf{Normalized DAN} performs better than \textbf{Diplomatic DAN} in most fields (``year of enrollment'', ``jurisdiction'', ``department'', ``date field 1'' and ``date field 2''), while \textbf{Diplomatic DAN} performs considerably better in fields that contain names and last names (``enrolee's full name'', ``1st parent's name'' and ``2nd parent's name''). We analyze the outputs based on this division of fields.

\medskip

\noindent
\textbf{Fields that do not contain names and last names}
Six out of the nine fields do not contain names or last names. In these cases, the results suggest that giving the model a normalized annotation to learn (which is what was used for Normalized DAN), improves the model's learning capabilities. The number of errors that Diplomatic DAN makes in all normalizable fields is shown in Table~\ref{tab:errors_normalized_fields}. 
Even if the results obtained by the two models can be comparable, training the \textbf{Normalized model} is less costly in terms of annotation effort and it also generates a formatted output, which is usually preferred when saving data into information systems. 

\medskip

\noindent
\textbf{Fields that contain names and last names}
As mentioned before, there are three fields in the dataset that contain names and last names: ``enrollee's full name'', ``1st parent's name'' and ``2nd parent's name''. In the three cases, the CER and WER, if compared ignoring cases and accents, are better with \textbf{Diplomatic DAN}. It is interesting to see how this improvement is different in the tree mentioned fields. 

The field ``enrollee's full name'' is only slightly better for Diplomatic DAN than for Normalized DAN. It is worth remembering that, in this specific field, where names and last names typically appear in several lines, in the margin of the certificate, a full annotation is provided both in the normalized dataset and in the diplomatic one. This probably explains the very similar results that are obtained by both models. 

The main difference between the ``1st parent's name'' and ``2nd parent's name'' fields is that the former is surrounded by preprinted text while the latter is always surrounded by handwritten text. This probably explains why the behavior of Diplomatic DAN in the ``2nd parent's name'' field is almost 3 times better than Normalized DAN, while the improvement obtained by Diplomatic DAN on the ``1st parent's field'' is more than 5 times better (when evaluated ignoring cases and accents). 

One of the possible reasons why Diplomatic DAN works better than Normalized DAN on these three fields was already explained in the description of the dataset: annotations provided by the civil registry included some abbreviation criteria on both parents' names fields where, very frequently, the second name was abbreviated to its initial (e.g., ``Alberto Carlos Bustos'' was transcribed as ``Alberto C. Bustos''). Somehow, this was learned by the model but with mixed results. In cases where the enrollee has a second name and the model was able to understand which letter to use, Normalized DAN produced a correct output (i.e. correctly abbreviated the middle name). In other cases, it made a mistake when abbreviating the middle name. In cases where no middle name was handwritten in the document, the model just invented one, presumably to agree with the format of the data it was trained on. 

\section{Conclusions and future work}\label{sec:conclusions} 

In this work we explored the capabilities of fine-tuning DAN using two annotation strategies to adapt it to a different context than the one in which it was trained. We observed that, even with very little data (181 documents), the model was able to obtain very similar results to the ones shown in the original paper. 

We also explored two annotation strategies, normalized and diplomatic, and saw that the former worked better in fields that can be standardized while the latter performed better in fields where character by character transcription is needed. The aforementioned results on fine-tuning DAN for transcribing handwritten documents open several questions.

The first one is whether using a ``hybrid'' annotated dataset (meaning, normalized for normalizable fields and diplomatic for fields containing names and last names) would result in a model that is capable of both transcribing ``normalizable'' fields in a normalized way (hence, eliminating the need of post-transcription normalization) while being extremely accurate when transcribing non-normalizable fields (such as, in our case, names and last names). 

A second natural line is exploring how far we can go in reducing the number of birth certificates needed to fine-tune a model without losing accuracy. 

A third line we will explore is the model's generalization capabilities. For instance, when dealing with different document layouts. Related to this, it is worth testing whether using a different annotation order for birth certificates that have the margin on the left or on the right will improve accuracy.

Further, one can explore whether exploiting document redundancy by annotating redundant information will result in higher accuracy. As can be seen from Figure~\ref{fig:crop_acta_anotada.png}, birth certificates generally have some data redundancy such as the enrolled person's name is written both in the margin and in the body of the document. Lastly, since handwriting styles can be so different among one another, we would like to explore whether training a model on a specific writer can lead to a better understanding of the writer's specific handwriting style and, therefore, a higher transcription accuracy.

\begin{credits}
\subsubsection{\ackname} 
The research that originated the results presented in this
publication was partly supported by the Agencia Nacional de Investigación e Innovación (ANII) and the France 2030 CollabNext project.
\end{credits}

%
%
\bibliographystyle{splncs04}
\bibliography{referencias}

\end{document}